\documentclass[letterpaper, 10 pt, conference]{ieeeconf}  

\IEEEoverridecommandlockouts                              

\usepackage{cite}
\usepackage{amsmath,amssymb,amsfonts}
\usepackage{algorithmic}
\usepackage{sidecap}
\usepackage{graphicx}
\usepackage{textcomp}
\usepackage{lipsum}  
\usepackage[font=footnotesize]{caption}
\usepackage{subcaption}
	
\usepackage[table]{xcolor}

\title{\LARGE \bf
Optical flow-based branch segmentation for complex orchard environments
}

\author{Alexander You$^1$, Cindy Grimm$^1$, Joseph R. Davidson$^1$
\thanks{This work has been submitted to the IEEE for possible publication. Copyright may be transferred without notice, after which this version may no longer be accessible.}
\thanks{This research is supported in part by USDA-NIFA through the Agriculture and Food Research Initiative, Agricultural Engineering Program (award No. 2020-67021-31958) and the AI Research Institutes program supported by NSF and USDA-NIFA under the AI Institute: Agricultural AI for Transforming Workforce and Decision Support (AgAID) (award No. 2021-67021-35344).}
\thanks{$^1$Collaborative Robotics and Intelligent Systems (CoRIS) Institute, Oregon State University, Corvallis OR 97331, USA {\tt\footnotesize \{youa, cindy.grimm, joseph.davidson\}@oregonstate.edu}}%
}

\begin{document}

\maketitle
\thispagestyle{empty}
\pagestyle{empty}

\begin{abstract}
Machine vision is a critical subsystem for enabling robots to be able to perform a variety of tasks in orchard environments. However, orchards are highly visually complex environments, and computer vision algorithms operating in them must be able to contend with variable lighting conditions and background noise. Past work on enabling deep learning algorithms to operate in these environments has typically required large amounts of hand-labeled data to train a deep neural network or physically controlling the conditions under which the environment is perceived. In this paper, we train a neural network system {\em in simulation only} using simulated RGB data and optical flow. This resulting neural network is able to perform foreground segmentation of branches in a busy orchard environment {\em without} additional real-world training or using any special setup or equipment beyond a standard camera. Our results show that our system is highly accurate and, when compared to a network using manually labeled RGBD data, achieves significantly more consistent and robust performance across environments that differ from the training set.
\end{abstract}

\section{Introduction}


Accurate sensing of the environment is an integral component of any robotic system. One common problem hindering many computer vision algorithms for specialty crop production (i.e. fresh market fruits and vegetables) is that outdoor environments --- like fruit orchards --- are very noisy with variable lighting conditions and highly complex and unstructured environments. In these environments, it is often useful to distinguish between parts of the image that are part of the foreground versus those in the background, i.e. to perform {\em foreground segmentation}. Fast and reliable foreground segmentation in orchard environments is directly applicable to a variety of orchard tasks (e.g. canopy estimation and blossom counting) and is also useful for many downstream 2.5D and 3D computer vision and control problems (e.g. localizing objects for pruning, thinning, and picking).

Foreground segmentation in orchards is a difficult problem, and because of this, many algorithms and research experiments are performed in highly controlled environments to minimize the effects of background noise. In our previous pruning experiments~\cite{you2021precision}, we trained a reinforcement learning system for dormant tree pruning using a computer vision system that, while successful, required the lab tree setup to be next to a blank wall to eliminate background noise. Similar examples can be found in many other works:~\cite{Tabb2017a} and~\cite{Lavaquiol2021} require backdrops to reconstruct tree models via photogrammetry, while~\cite{Botterill2017c} uses a trailer that encompasses the target grape vine, isolating it in the scene from the adjacent vineyard rows and enabling complete control of the, essentially, indoor lighting conditions. Deep learning using depth data is also a common approach. However, training these networks typically requires a large amount of hand-labeled data, and the resulting networks can struggle with environmental conditions that differ even slightly from the training data. 

\begin{figure}[bt]
	\centering
\includegraphics[width = \columnwidth]{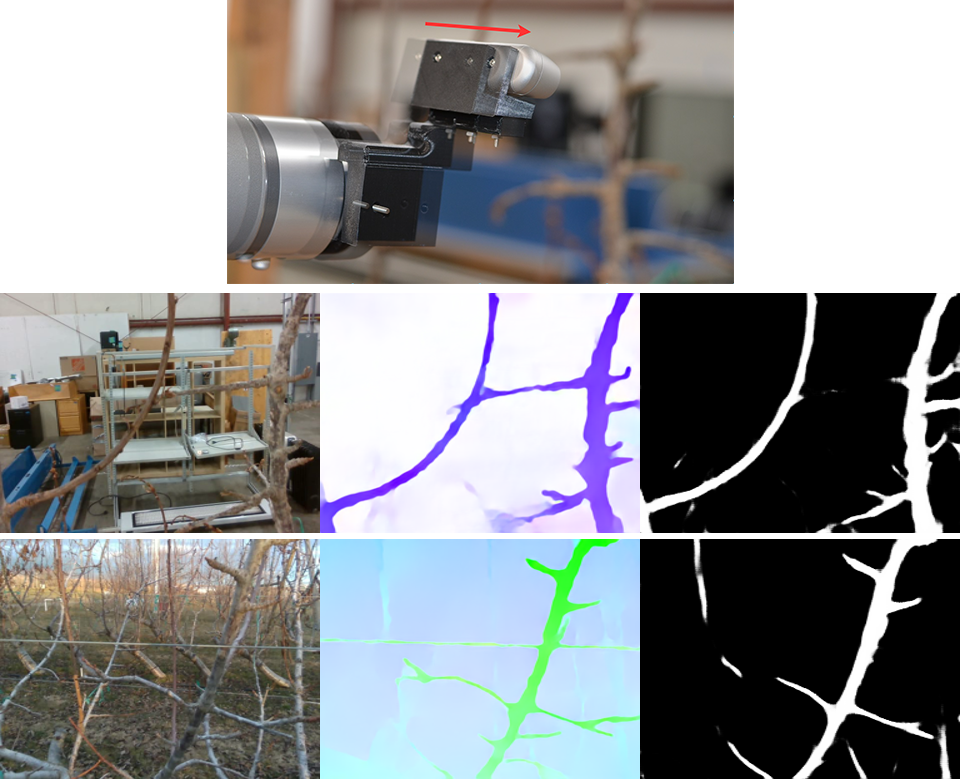}\caption{Our system uses movement from a mobile camera, such as an eye-in-hand robotic configuration, to yield dense optical flow estimates (middle), allowing us to perform robust foreground branch segmentation (right).}\label{fig:abstract}
    \vspace{-2ex}
\end{figure}

As humans, one way we contend with the issue of background noise is to move our heads to get different views of the scene. Our goal is to apply this principle to a camera mounted in an eye-in-hand configuration on a robotic manipulator, such as shown in Figure~\ref{fig:abstract}. When moving a camera translationally, due to parallax, objects in the foreground will move more than those in the background. Such disparities are the foundation of stereo matching algorithms that reconstruct 3D models from image pairs. However, a full 3D reconstruction is generally unnecessary for many orchard applications and requires precise knowledge of the camera's calibration parameters and pose change. Instead, we compute the dense \textit{optical flow}, i.e. movement vectors for each pixel in the image estimating their speeds across frames. This optical flow serves as a proxy for depth data, enabling us to do real-time foreground object segmentation.

In this paper, we introduce a deep neural network for producing foreground masks for trees in planar orchard environments by augmenting RGB data with optical flow data computed from the previous RGB frame. The augmented image with RGB and optical flow is fed into the neural network to produce a segmentation mask. To our knowledge, our approach is the first one for segmentation of branches in outdoor environments that is trained entirely in simulation, as well as the first to make use of optical flow data instead of relying on depth data. By training our network this way, we eliminate the need to do any manual labeling of ground truth data. We demonstrate that both of these components combined allow for consistent and accurate segmentation results by evaluating the performance of a suite of comparison networks in four outdoor environments and one indoor environment and showing that only our system was able to perform consistently across all of them. 

\section{Related Work}









Branch detection is an active area of research within agricultural robotics. For this paper, we focus only on techniques used to perform segmentation on 2D images, rather than techniques for 3D reconstruction of trees. Many papers have shown success in using machine learning techniques to create branch masks. Amatya and Karkee~\cite{Amatya2016} train a Bayesian classifier to segment and reconstruct branches on images of sweet cherry trees taken at night, while~\cite{Song2021} uses an encoder-decoder network to segment out branches, wires, and fruit in a kiwifruit orchard. Depth information is a common modality for filtering out unwanted background noise: ~\cite{Majeed2020,Zhang2021} use depth information to filter out all points beyond a specified threshold before feeding the RGB image through a neural network, while~\cite{Zhang2018,Chen2021} feed the depth channel directly into the neural network to filter out background noise. Yang et al.~\cite{Yang2020} produce a mask with a raw RGB image but post-process the mask with depth data to create an accurate tree model for localization. Depth data has a number of notable limitations, particularly that the quality and characteristics of depth data can vary greatly between different depth cameras. Because of this, realistic depth images are difficult to simulate, requiring training data to be labeled manually, a very time-consuming and arduous process. Our work substitutes optical flow as a proxy for depth data, allowing us to train a network using only simulated images and transfer the resulting network to real data.

Our problem is closely related to the issue of \textit{background subtraction} (or, equivalently, foreground segmentation) in which the goal is to use a stream of images to detect and extract objects of interest in a scene by removing pixels which are located in the background. Background subtraction is a highly studied area of computer vision, with numerous surveillance-related applications as detailed by~\cite{GarciaGarcia2020}. Piccardi~\cite{Piccardi2004} offers a survey of classical background subtraction techniques, which typically operate by constructing a model of the background and maintaining it against the most current frame. Bouwmans et al.~\cite{Bouwmans2019} review the large body of research that has shown success with using convolutional neural networks to perform foreground segmentation, often with robustness to issues such as camera jitter, dynamic backgrounds, and weather conditions that traditional methods struggled with. However, in addition to these networks being highly scene-specific, the vast majority of background subtraction algorithms assume the camera is mostly stationary and the object of interest is moving, whereas in our situation, all objects of interest are stationary while the camera itself is moving. 

Our research is also closely related to the areas of object tracking and mask propagation for video data. Yao et al.~\cite{Yao2020} gives a comprehensive overview of many of these methods. Of particular interest to us is the use of deep neural networks such as SegFlow~\cite{Cheng2017} that simultaneously learn optical flow and segmentation data, ultimately resulting in improved object segmentations when compared to learning them individually. Though we do not train our network in the same way as theirs, we follow a similar idea by using motion induced by camera movement to segment out the objects of interest in the foreground.

\section{Problem Statement}

\begin{figure}[bt]
	\centering
\includegraphics[width = \columnwidth]{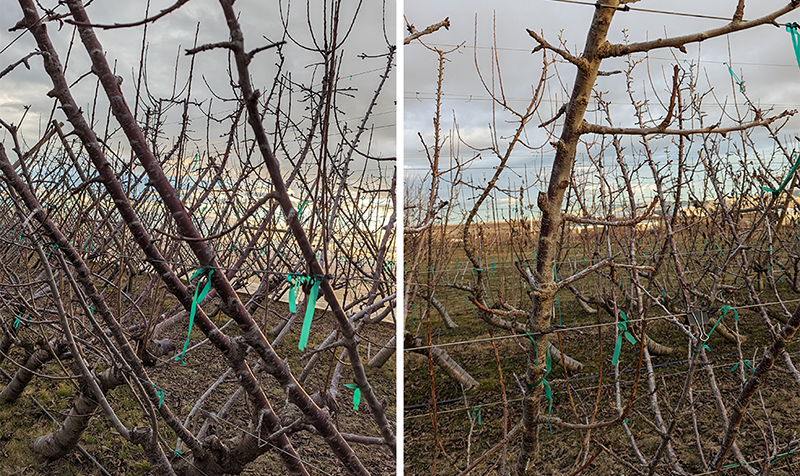}\caption{Upright fruiting offshoot cherry trees in a V-trellis orchard after leaf drop. The repetitive nature of the orchards makes it difficult to distinguish foreground branches from the background using a single RGB image.}\label{fig:env}
    \vspace{-2ex}
\end{figure}

Our work focuses primarily on modern, high-density orchard systems, such as the one shown in Figure~\ref{fig:env}. This Figure shows sweet cherry trees (during the dormant season) trained in an upright fruiting offshoot (UFO) system whereby vertical leader branches growing from the nearly horizontal trunk are tied to trellis wires, creating an inclined, V-style planar canopy. Our goal is to generate a mask that will highlight only the branches attached to the front plane of the trellis, filtering out all other objects including the ground, sky, trees and objects in different rows, trellis wires, and tags and non-organic parts of the foreground.

\begin{figure*}[bt]
	\centering
\includegraphics[width = 0.7\textwidth]{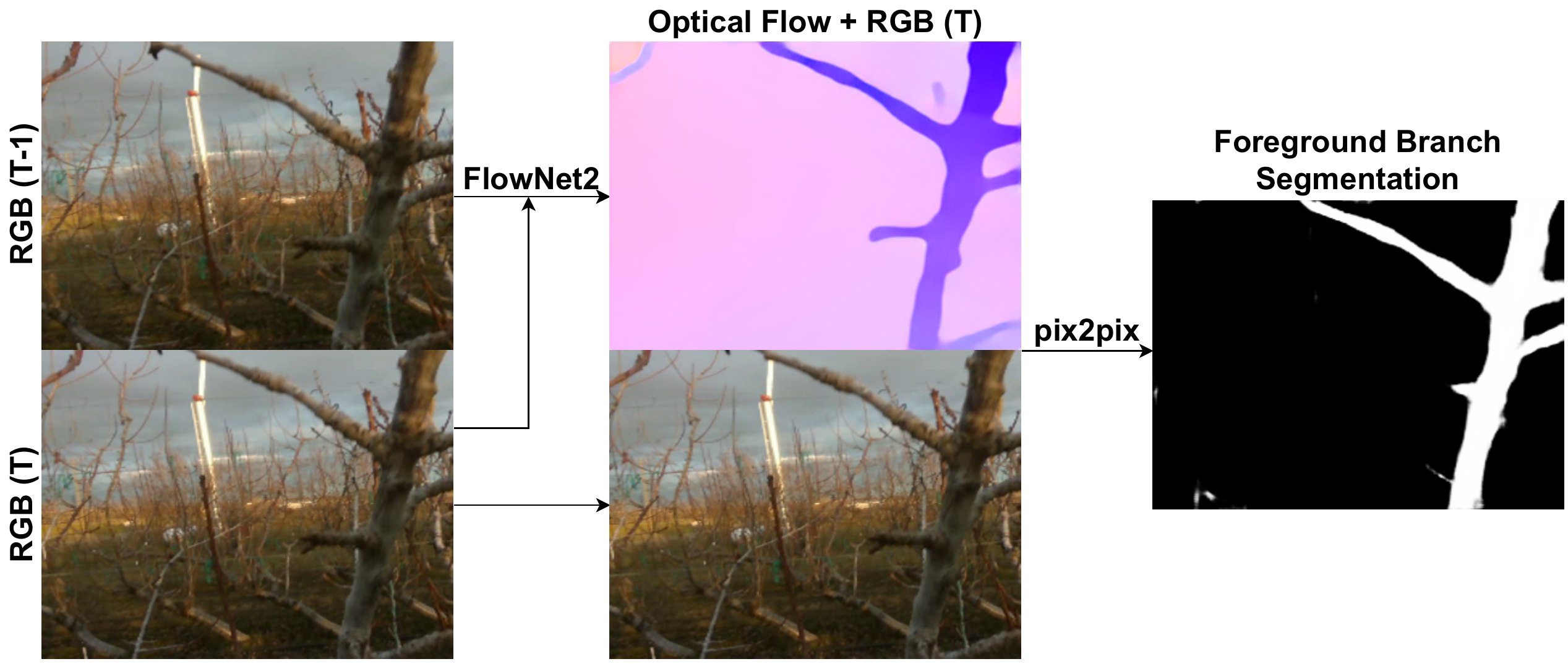}\caption{Our segmentation framework. We start by computing optical flow using two RGB frames created by a small camera pose change. Afterwards, we stack the optical flow with the RGB image and feed it into the pix2pix GAN, which performs the segmentation.}\label{fig:system}
    \vspace{-2ex}
\end{figure*}

We assume that our setup for operating in these environments consists of a mobile camera oriented roughly parallel to the planar trellis system. We also assume that the trees themselves are mostly stationary (no excessive wind). Therefore, when the camera moves translationally, the objects in the foreground will move the most.

\section{Optical Flow-Based Network Details}
\label{sec:flownetwork}

In this section, we describe the architecture and training methodology of our segmentation network, which combines an existing optical flow network (FlowNet2) with the pix2pix GAN (Section~\ref{sec:networks:ours}). Notably, we train our network using entirely simulated data (Section~\ref{sec:simdata}), avoiding the need to label data by hand.

\subsection{Architecture}
\label{sec:networks:ours}

Our segmentation network is shown in Figure~\ref{fig:system}. We utilize the pix2pix~\cite{pix2pix2017} framework, a Generative Adversarial Network (GAN)~\cite{goodfellow2014generative} designed to convert images in one domain into another one, to perform the segmentation. Our previous work~\cite{you2021precision}, as well as others~\cite{Chen2021}, have shown that the pix2pix network is capable of performing robust segmentation. 

In addition to passing the 3-channel RGB image into the GAN, which outputs a 1-channel segmentation mask, we also pass in a 3-channel colorized representation of the optical flow between the current RGB frame and a previous one. To compute the optical flows, we use a fully-sized FlowNet2 network with pre-trained weights~\cite{Ilg2017}, which we chose due to the high quality of the optical flows produced as well as its ability to work on simulated images and on image pairs with varying distances. 

The combined network's runtime is nearly real-time, mainly limited by the performance of FlowNet2. On our research computer using an NVIDIA GeForce GTX 1050 Ti, we observed runtimes of the pix2pix network to be around 20 FPS and FlowNet2 to be around 4 FPS, though the original paper for FlowNet2 purports to reach 8 FPS using an NVIDIA GTX 1080.

\subsection{Simulated Image Generation}
\label{sec:simdata}

\begin{figure}[bt]
	\centering
\includegraphics[width = \columnwidth]{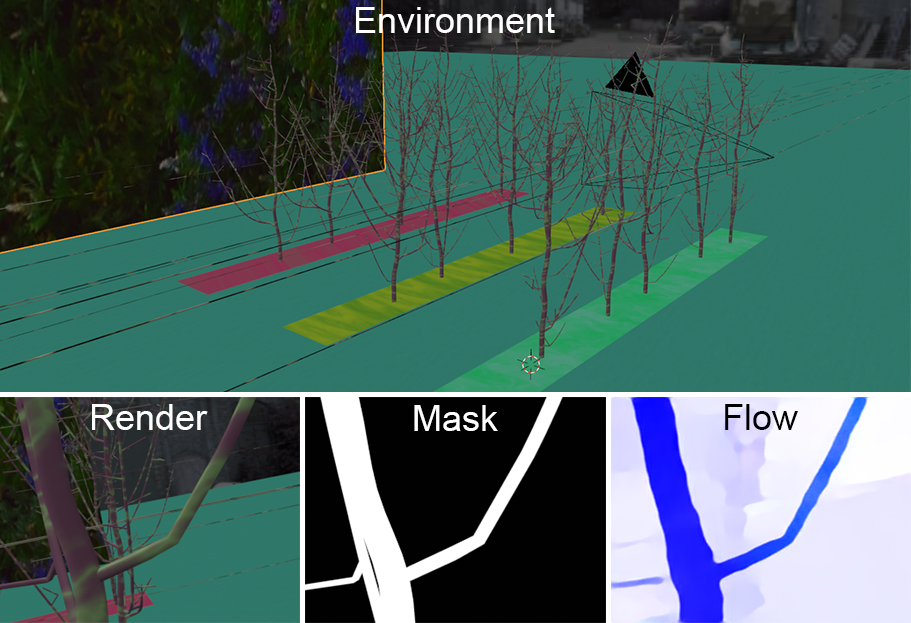}\caption{A simulated orchard scene with randomized textures in Blender, along with the corresponding render, foreground branch mask, and optical flow computed using a second frame.}\label{fig:simenv}
    \vspace{-2ex}
\end{figure}

To create a network that can generalize to real world scenarios without requiring any manual labeling of data, we follow the methodology of~\cite{James2019}. This approach trains a GAN to convert an image of a scene to a simplified, canonical representation (essentially a labeled mask). The key to this approach is training the GAN using simulated images with heavy texture and lighting randomization. They showed that the resulting GAN was able to generalize surprisingly well to real images of the scene. We showed in our previous work~\cite{you2021precision} that this process also worked for tree branches in a simple lab setup. This paper further augments the process with optical flow data to allow our system to work in complicated environments in which a single RGB image alone is not enough to perform a proper segmentation.

For our experiments, we created a multi-row orchard environment in Blender, an open source 3D modeling software. A typical environment with randomized textures is shown in Figure~\ref{fig:simenv}. Each environment was created by randomly lining up 1-5 rows of tree models, where the spacing between trees was about 0.6m and the spacing between rows was about 1.1m. These values were based on measurements from a commercial UFO cherry tree orchard. Each row of trees has 3 trellis wires placed at random vertical offsets between 0-1m above the ground, as well as a rectangular plane on the ground representing the soil area. We also placed between 0 and 2 randomly sized rectangular walls behind the final tree row to simulate nearby walls or foreign objects.

Once the environment is generated, we randomize the textures of the trees (each tree receives the same texture), ground plane, soil area planes, and trellis wires by picking textures from a set of 5000 simple textures. Each of the walls was textured with one of 1000 random images sourced from the Open Images v6 data set~\cite{kuznetsova2020open}, and the lighting of the scene was determined using one of 303 randomly chosen high-dynamic-range images (HDRIs). For all textures, we randomly shifted the hue, saturation, and values of the source images to obtain a wide variety of synthetic environments.

After randomizing the textures as described above, we randomly place a camera so that it faces the front row of the scene and apply a random angular perturbation of up to 15 degrees. We then generated two renders of the scene by moving the camera a random direction by 0.5-2 cm. For each render, we created a corresponding ground truth mask by coloring the foreground trees white, hiding all other objects, and setting the background to black.

Each pair of renders is used to generate a corresponding pair of colorized optical flow images from FlowNet2, giving us two sets of corresponding synthetic images and optical flow images. For our training, we generate a total of 4400 6-channel synthetic-flow pairs, 4000 of which are used for the training phase and 400 of which are used for validation; training details are given in Section~\ref{sec:networks}. Figure~\ref{fig:simenv} shows an example of a render, mask, and optical flow image generated from a synthetic pair.

\subsection{Training Methodology}
\label{sec:networks}

After precomputing the optical flows for each pair of renders, we trained the pix2pix network using the synthetic-flow pairs. Since the optical flows are precomputed, the FlowNet2 network weights are not updated during training, decoupling the optical flow computation from the segmentation.

We trained the network for a maximum of 100000 images. Every 1000 images, we computed the average L1 loss for the validation data set and saved the network weights if the average loss was lower than the previous best. Although in general for GANs L1 loss is not the best criterion for determining network convergence, for this relatively simple segmentation task we would expect that good segmentations would be highly correlated with low L1 losses.

\section{Comparison Network Details}
\label{sec:networks_other}

In this section, we cover the real world data set we collected (Section~\ref{sec:realdata}) that we used to train a number of comparison networks (Section~\ref{sec:otherdata}). The networks trained using this data serve as a useful comparison baseline for our own network trained using synthetic data.

\subsection{Real World Training Data Set}
\label{sec:realdata}

\begin{figure}[bt]
	\centering
\includegraphics[width = \columnwidth]{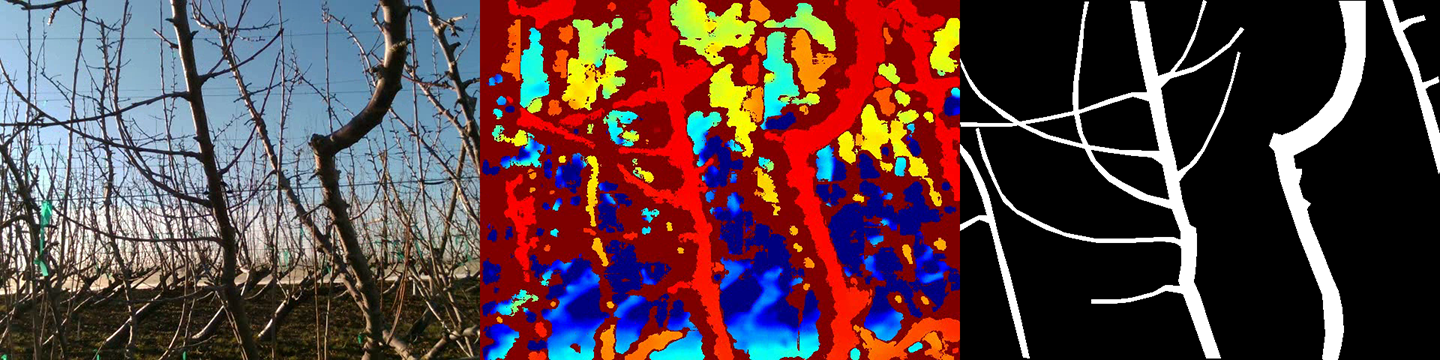}\caption{A sample RGB, colorized depth, and labeled ground truth image from the training set.}\label{fig:trainimg}
    \vspace{-2ex}
\end{figure}

To train a comparison network with hand-labeled data, we collected video RGB and depth data from a UFO cherry tree orchard on a sunny day at 1 PM in Prosser, Washington using an Intel (Santa Clara, CA, USA) Realsense D435 camera at a resolution of 640 x 480 pixels. A human operator collected the data by holding the camera in their hand and moving it in primarily translational directions at a distance of roughly 0.3-1m from the plane of trees. For ease of data collection, the depth data was colorized using a color map and appended to the RGB video.

After the data collection, we chose a total of 326 training images and 45 validation images from the videos to manually annotate with ground truth data. An example of RGBD and ground truth data is shown in Figure~\ref{fig:trainimg}. The labeling process was arduous, as due to the complexity of each image and the necessity to cross reference the source video to determine foreground movement, each image could take up to 10 minutes to label, resulting in a combined effort of over 40 hours just to label 371 images. Furthermore, we expect that there was some amount of label noise introduced by the complexity of the scene as well as ambiguities about whether certain branches qualified as foreground branches or not (e.g. a branch extending from the front trellis wall into the background).

\subsection{List of Comparison Networks}
\label{sec:otherdata}

Our goal is to evaluate whether our system --- using synthetic data only --- obtains comparable accuracy to a system trained using hand-labeled RGBD data while also being robust to a wide variety of environments. All trained networks use the pix2pix architecture and training methodology as described in Sections~\ref{sec:networks:ours} and~\ref{sec:networks}, but with the network inputs changed:

\begin{itemize}
    \item RGB data: Do we use real RGB data with labeled ground truths, or do we use synthetic data?
    \item Depth/flow: To assist with depth perception, do we use depth data, computed optical flow, or nothing at all (relying on a single frame of RGB data)?
\end{itemize}

\noindent Our baseline networks are as follows:

\begin{itemize}
    \item Synthetic with optical flow (Syn+F*), 6 channels: Our proposed network (Section~\ref{sec:flownetwork}) trained using the synthetic image-colorized flow image pairs described in Section~\ref{sec:simdata}.
    \item RGB with colorized depth (RGB+D), 6 channels: Trained using the manually labeled RGBD data described in Section~\ref{sec:realdata}. This setup is comparable to the one described in \cite{Chen2021}.
\end{itemize}

To further examine the effects of the use of depth and flow data, we train three other networks for an ablation study:

\begin{itemize}
    \item Synthetic with no flow (Syn), 3 channels: Trained using the synthetically generated images, but without including the colorized optical flow image. 
    \item RGB with no depth (RGB), 3 channels: Trained using the manually labelled data set, but without including the colorized depth image.
    \item RGB with optical flow (RGB+F), 6 channels: Trained using the RGB images collected from the main data set, but instead of using the depth image, we use the colorized flow associated with either the frame before or the frame after the image. Since there are two optical flow images per RGB image, this doubles the sizes of the training and validation sets. 
\end{itemize}

\section{Experiments}
\label{sec:experiments}

In this section we discuss the various test data sets we use to evaluate the comparison networks (Section~\ref{sec:testimages}), as well as the metrics used to perform the evaluation (Section~\ref{sec:metrics}). In addition to assessing each network's relative performance against our own (Syn+F*), our goal is to examine how each network's performance degrades as the data set changes. These results are covered in Section~\ref{sec:results}.

\subsection{Test Image Data Sets}
\label{sec:testimages}

To evaluate the robustness of the trained networks, in addition to the training and validation data from the sweet cherry tree orchard described in Section~\ref{sec:realdata}, we labeled 15 ground truth images each from 5 different environments using the same data collection process as in Section~\ref{sec:realdata}. The environments, shown in Figure~\ref{fig:results}, are listed here in order of their similarity to the training data set:

\begin{enumerate}
    \item Main: Separate data collected on the same day and at the same time as the training data. 
    \item Afternoon: Data collected from the same orchard with sunny conditions on a different day, but at 4 PM instead of 1 PM. Notably, the sun was setting, resulting in image glare and exaggerated contrast.
    \item Cloudy: Data collected from the same orchard, but under mostly cloudy conditions, at 5 PM on a different day.
    \item Envy spindle trees, 12 PM, cloudy: Data collected from a different orchard of Envy apples grown in a tall, spindle configuration. The day was foggy and cloudy. These images differ visually from the main data set both in the weather conditions and the general appearance of the trees.
    \item Lab setup: Data collected in our lab using our experimental setup from our previous work~\cite{you2021precision} which uses branches from cherry trees. For this setup, the camera was mounted onto a Universal Robots (Odense, Denmark) UR5e robot and manually moved using the control panel as the video data was collected. We have moved the experimental setup away from a blank wall so that a large amount of background clutter is present.
\end{enumerate}

\subsection{Evaluation metrics}
\label{sec:metrics}

To measure the quality of the prediction masks, we first create a binary prediction mask by thresholding all predictions (which range from 0 to 1) that are below 0.5 to 0 and all predictions above 0.5 to 1. We denote the set of all positively identified pixels in the ground truth and predictions as $T$ and $P$ and all negative pixels as $T^-$ and $P^-$ respectively. We then use the following standard image segmentation metrics to evaluate the performance:

\begin{itemize}
    \item Intersection over union (IOU): $IOU = \frac{\left|T \cap P\right|}{\left|T \cup P\right|} \in [0, 1]$
    \item False-positive rate, relative to image size (FP Rate): $FP = \frac{\left|T^- \cap P\right|}{\left|T \cup T^-\right|} \in [0, 1]$
    \item False-negative rate, relative to ground truth size (FN Rate): $FN = \frac{\left|T \cap P^-\right|}{\left|T\right|} \in [0, 1]$
\end{itemize}

\section{Results}
\label{sec:results}

\begin{figure*}[bt]
	\centering
\includegraphics[width = \linewidth]{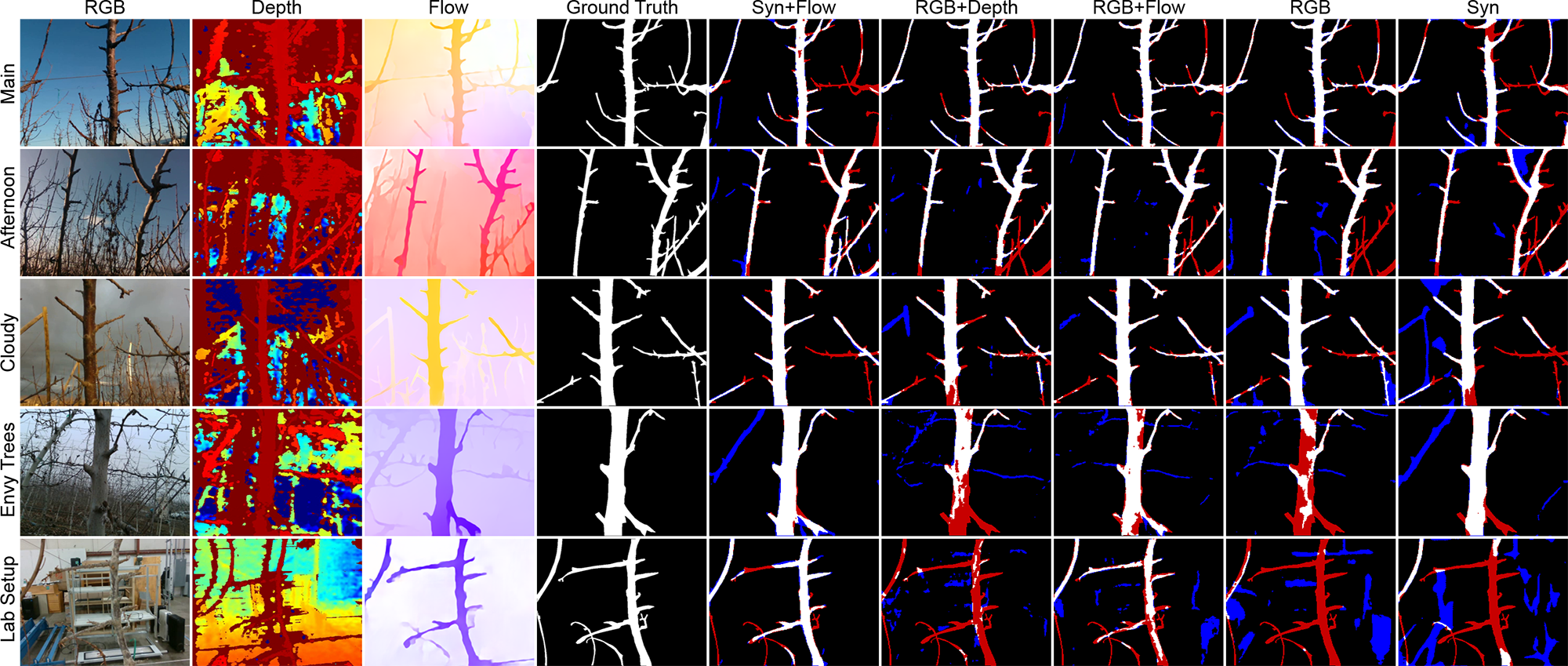}\caption{Examples of images from the five test data sets (Section~\ref{sec:testimages}), as well as the resulting segmentations from each trained network (Section~\ref{sec:networks}). For the segmentations, red pixels represent false negatives, and blue pixels represent false positives.}\label{fig:results}
    \vspace{-2ex}
\end{figure*}

\begin{table}
\centering
\caption{Averaged results for three metrics across all networks and data sets. Our network (Syn+F*) has the most consistent performance across all data sets for each metric.}
\begin{tabular}{|c|ccccc|}
\hline
{\textbf{IOU}} & {Main} & {Afternoon} & {Cloudy} & {Envy} & {Lab} \\
\textbf{Syn+F*} & {\cellcolor[HTML]{92D28F}} \color[HTML]{000000} 62.3\% & {\cellcolor[HTML]{A7DBA0}} \color[HTML]{000000} 53.3\% & {\cellcolor[HTML]{81CA81}} \color[HTML]{000000} 69.5\% & {\cellcolor[HTML]{A7DBA0}} \color[HTML]{000000} 53.7\% & {\cellcolor[HTML]{98D594}} \color[HTML]{000000} 60.1\% \\
\textbf{RGB+D} & {\cellcolor[HTML]{7DC87E}} \color[HTML]{000000} 71.3\% & {\cellcolor[HTML]{AFDFA8}} \color[HTML]{000000} 49.5\% & {\cellcolor[HTML]{AFDFA8}} \color[HTML]{000000} 49.7\% & {\cellcolor[HTML]{C7E9C0}} \color[HTML]{000000} 37.9\% & {\cellcolor[HTML]{F4FBF2}} \color[HTML]{000000} 3.2\% \\
RGB+F & {\cellcolor[HTML]{75C477}} \color[HTML]{000000} 74.8\% & {\cellcolor[HTML]{88CE87}} \color[HTML]{000000} 66.6\% & {\cellcolor[HTML]{7AC77B}} \color[HTML]{000000} 72.5\% & {\cellcolor[HTML]{AADDA4}} \color[HTML]{000000} 52.1\% & {\cellcolor[HTML]{CEECC8}} \color[HTML]{000000} 33.3\% \\
RGB & {\cellcolor[HTML]{97D492}} \color[HTML]{000000} 60.7\% & {\cellcolor[HTML]{ACDEA6}} \color[HTML]{000000} 50.7\% & {\cellcolor[HTML]{9ED798}} \color[HTML]{000000} 57.9\% & {\cellcolor[HTML]{D9F0D3}} \color[HTML]{000000} 26.4\% & {\cellcolor[HTML]{F5FBF2}} \color[HTML]{000000} 2.9\% \\
Syn & {\cellcolor[HTML]{A5DB9F}} \color[HTML]{000000} 54.5\% & {\cellcolor[HTML]{C2E7BB}} \color[HTML]{000000} 40.4\% & {\cellcolor[HTML]{9BD696}} \color[HTML]{000000} 58.6\% & {\cellcolor[HTML]{B6E2AF}} \color[HTML]{000000} 45.7\% & {\cellcolor[HTML]{E5F5E1}} \color[HTML]{000000} 18.7\% \\ \hline
{\textbf{FN Rate}} & {Main} & {Afternoon} & {Cloudy} & {Envy} & {Lab} \\
\textbf{Syn+F*} & {\cellcolor[HTML]{FEDBCC}} \color[HTML]{000000} 28.7\% & {\cellcolor[HTML]{FDD4C2}} \color[HTML]{000000} 33.2\% & {\cellcolor[HTML]{FEE3D6}} \color[HTML]{000000} 22.2\% & {\cellcolor[HTML]{FED8C7}} \color[HTML]{000000} 30.7\% & {\cellcolor[HTML]{FEDFD0}} \color[HTML]{000000} 26.1\% \\
\textbf{RGB+D} & {\cellcolor[HTML]{FEE3D7}} \color[HTML]{000000} 21.8\% & {\cellcolor[HTML]{FDC6B0}} \color[HTML]{000000} 42.2\% & {\cellcolor[HTML]{FDC6B0}} \color[HTML]{000000} 42.2\% & {\cellcolor[HTML]{FCBDA4}} \color[HTML]{000000} 48.6\% & {\cellcolor[HTML]{FB7151}} \color[HTML]{F1F1F1} 95.5\% \\
RGB+F & {\cellcolor[HTML]{FEE7DB}} \color[HTML]{000000} 17.7\% & {\cellcolor[HTML]{FEE1D4}} \color[HTML]{000000} 23.8\% & {\cellcolor[HTML]{FEE6DA}} \color[HTML]{000000} 18.1\% & {\cellcolor[HTML]{FDD4C2}} \color[HTML]{000000} 33.0\% & {\cellcolor[HTML]{FCB69B}} \color[HTML]{000000} 53.1\% \\
RGB & {\cellcolor[HTML]{FDD7C6}} \color[HTML]{000000} 31.7\% & {\cellcolor[HTML]{FDD1BE}} \color[HTML]{000000} 35.3\% & {\cellcolor[HTML]{FED9C9}} \color[HTML]{000000} 29.9\% & {\cellcolor[HTML]{FCA486}} \color[HTML]{000000} 64.5\% & {\cellcolor[HTML]{FB7353}} \color[HTML]{F1F1F1} 94.2\% \\
Syn & {\cellcolor[HTML]{FED8C7}} \color[HTML]{000000} 30.8\% & {\cellcolor[HTML]{FDC5AE}} \color[HTML]{000000} 43.4\% & {\cellcolor[HTML]{FEDFD0}} \color[HTML]{000000} 26.2\% & {\cellcolor[HTML]{FDCDB9}} \color[HTML]{000000} 37.8\% & {\cellcolor[HTML]{FCAE92}} \color[HTML]{000000} 58.6\% \\ \hline
{\textbf{FP Rate}} & {Main} & {Afternoon} & {Cloudy} & {Envy} & {Lab} \\
\textbf{Syn+F*} & {\cellcolor[HTML]{FEE6DA}} \color[HTML]{000000} 1.8\% & {\cellcolor[HTML]{FDD7C6}} \color[HTML]{000000} 3.1\% & {\cellcolor[HTML]{FEE7DC}} \color[HTML]{000000} 1.7\% & {\cellcolor[HTML]{FDD5C4}} \color[HTML]{000000} 3.3\% & {\cellcolor[HTML]{FEE2D5}} \color[HTML]{000000} 2.3\% \\
\textbf{RGB+D} & {\cellcolor[HTML]{FFEBE2}} \color[HTML]{000000} 1.2\% & {\cellcolor[HTML]{FEE7DC}} \color[HTML]{000000} 1.7\% & {\cellcolor[HTML]{FEE5D9}} \color[HTML]{000000} 2.0\% & {\cellcolor[HTML]{FED8C7}} \color[HTML]{000000} 3.1\% & {\cellcolor[HTML]{FDCCB8}} \color[HTML]{000000} 3.9\% \\
RGB+F & {\cellcolor[HTML]{FEEAE1}} \color[HTML]{000000} 1.3\% & {\cellcolor[HTML]{FEE9DF}} \color[HTML]{000000} 1.5\% & {\cellcolor[HTML]{FEE8DE}} \color[HTML]{000000} 1.5\% & {\cellcolor[HTML]{FEDFD0}} \color[HTML]{000000} 2.6\% & {\cellcolor[HTML]{FDC9B3}} \color[HTML]{000000} 4.1\% \\
RGB & {\cellcolor[HTML]{FEE8DD}} \color[HTML]{000000} 1.6\% & {\cellcolor[HTML]{FEDECF}} \color[HTML]{000000} 2.7\% & {\cellcolor[HTML]{FEDECF}} \color[HTML]{000000} 2.7\% & {\cellcolor[HTML]{FDD5C4}} \color[HTML]{000000} 3.3\% & {\cellcolor[HTML]{FA6547}} \color[HTML]{F1F1F1} 10.2\% \\
Syn & {\cellcolor[HTML]{FDD3C1}} \color[HTML]{000000} 3.4\% & {\cellcolor[HTML]{FCBEA5}} \color[HTML]{000000} 4.8\% & {\cellcolor[HTML]{FDD4C2}} \color[HTML]{000000} 3.3\% & {\cellcolor[HTML]{FDC7B2}} \color[HTML]{000000} 4.2\% & {\cellcolor[HTML]{F44D38}} \color[HTML]{F1F1F1} 11.5\% \\ \hline
\end{tabular}
\label{table:stats}
\end{table}

\begin{table}
\centering
\caption{$p$-values for a Welch's $t$-test comparing IOUs. The top table compares the performance of other networks against our own (Syn+F*) on each data set, while the bottom table examines performance degradation for each network against the Main data set. (Colors: Green if IOU is significantly higher than baseline, red if significantly lower; $p$-values below 0.05, 0.01 and 0.001 highlighted)}
\begin{tabular}{|c|ccccc|} \hline
{\textbf{Vs. Syn+F}} & {Main} & {Afternoon} & {Cloudy} & {Envy} & {Lab} \\
\textbf{Syn+F*} & {\cellcolor[HTML]{F7F7F6}} \color[HTML]{000000} $\downarrow$ & {\cellcolor[HTML]{F7F7F6}} \color[HTML]{000000} $\downarrow$& {\cellcolor[HTML]{F7F7F6}} \color[HTML]{000000} $\downarrow$ & {\cellcolor[HTML]{F7F7F6}} \color[HTML]{000000} $\downarrow$ & {\cellcolor[HTML]{F7F7F6}} \color[HTML]{000000} $\downarrow$ \\
\textbf{RGB+D} & {\cellcolor[HTML]{B7E085}} \color[HTML]{000000} 0.004 & {\cellcolor[HTML]{F7F7F6}} \color[HTML]{000000} 0.350 & {\cellcolor[HTML]{FCA486}} \color[HTML]{000000} 0.008 & {\cellcolor[HTML]{FCA486}} \color[HTML]{000000} 0.005 & {\cellcolor[HTML]{F44D38}} \color[HTML]{F1F1F1} 0.000 \\
RGB+F & {\cellcolor[HTML]{7FBC41}} \color[HTML]{000000} 0.000 & {\cellcolor[HTML]{7FBC41}} \color[HTML]{000000} 0.000 & {\cellcolor[HTML]{F7F7F6}} \color[HTML]{000000} 0.451 & {\cellcolor[HTML]{F7F7F6}} \color[HTML]{000000} 0.740 & {\cellcolor[HTML]{F44D38}} \color[HTML]{F1F1F1} 0.000 \\
RGB & {\cellcolor[HTML]{F7F7F6}} \color[HTML]{000000} 0.746 & {\cellcolor[HTML]{F7F7F6}} \color[HTML]{000000} 0.516 & {\cellcolor[HTML]{F7F7F6}} \color[HTML]{000000} 0.079 & {\cellcolor[HTML]{F44D38}} \color[HTML]{F1F1F1} 0.000 & {\cellcolor[HTML]{F44D38}} \color[HTML]{F1F1F1} 0.000 \\
Syn & {\cellcolor[HTML]{FCA486}} \color[HTML]{000000} 0.009 & {\cellcolor[HTML]{FCA486}} \color[HTML]{000000} 0.005 & {\cellcolor[HTML]{FCBEA5}} \color[HTML]{000000} 0.034 & {\cellcolor[HTML]{F7F7F6}} \color[HTML]{000000} 0.199 & {\cellcolor[HTML]{F44D38}} \color[HTML]{F1F1F1} 0.000 \\ \hline
{\textbf{Vs. Main}} & {Main} & {Afternoon} & {Cloudy} & {Envy} & {Lab} \\
\textbf{Syn+F*} & {\cellcolor[HTML]{F7F7F6}} \color[HTML]{000000} $\longrightarrow$ & {\cellcolor[HTML]{FCA486}} \color[HTML]{000000} 0.009 & {\cellcolor[HTML]{F7F7F6}} \color[HTML]{000000} 0.086 & {\cellcolor[HTML]{F7F7F6}} \color[HTML]{000000} 0.061 & {\cellcolor[HTML]{F7F7F6}} \color[HTML]{000000} 0.493 \\
\textbf{RGB+D} & {\cellcolor[HTML]{F7F7F6}} \color[HTML]{000000} $\longrightarrow$ & {\cellcolor[HTML]{F44D38}} \color[HTML]{F1F1F1} 0.000 & {\cellcolor[HTML]{FCA486}} \color[HTML]{000000} 0.003 & {\cellcolor[HTML]{F44D38}} \color[HTML]{F1F1F1} 0.000 & {\cellcolor[HTML]{F44D38}} \color[HTML]{F1F1F1} 0.000 \\
RGB+F & {\cellcolor[HTML]{F7F7F6}} \color[HTML]{000000} $\longrightarrow$ & {\cellcolor[HTML]{F44D38}} \color[HTML]{F1F1F1} 0.000 & {\cellcolor[HTML]{F7F7F6}} \color[HTML]{000000} 0.338 & {\cellcolor[HTML]{F44D38}} \color[HTML]{F1F1F1} 0.000 & {\cellcolor[HTML]{F44D38}} \color[HTML]{F1F1F1} 0.000 \\
RGB & {\cellcolor[HTML]{F7F7F6}} \color[HTML]{000000} $\longrightarrow$ & {\cellcolor[HTML]{F7F7F6}} \color[HTML]{000000} 0.069 & {\cellcolor[HTML]{F7F7F6}} \color[HTML]{000000} 0.681 & {\cellcolor[HTML]{F44D38}} \color[HTML]{F1F1F1} 0.000 & {\cellcolor[HTML]{F44D38}} \color[HTML]{F1F1F1} 0.000 \\
Syn & {\cellcolor[HTML]{F7F7F6}} \color[HTML]{000000} $\longrightarrow$ & {\cellcolor[HTML]{FCA486}} \color[HTML]{000000} 0.002 & {\cellcolor[HTML]{F7F7F6}} \color[HTML]{000000} 0.307 & {\cellcolor[HTML]{F7F7F6}} \color[HTML]{000000} 0.101 & {\cellcolor[HTML]{F44D38}} \color[HTML]{F1F1F1} 0.000 \\ \hline
\end{tabular}
\label{table:pvals}
\end{table}

The averaged metrics for each network-environment pairing are shown in Table~\ref{table:stats}. Table~\ref{table:pvals} shows $p$-values for measuring if other networks had better or worse IOUs than our network (Syn+F*), as well as if networks had notable performance degradations against the Main data set. Examples of some of the segmentation outputs can be seen in Figure~\ref{fig:results}. In general, the results were in line with our expectations in that Syn+F* had the most consistent performance across all data sets, whereas all networks trained using RGB data began to degrade in performance as the environment began to differ from the training data set.

The networks trained without flow or depth (RGB and Syn), using only a single RGB image as input, generally had higher false positive rates and worse performance than the corresponding versions trained with depth or optical flow data, indicating that adding these features makes it significantly easier to discern the foreground from the background. This was consistent with the annotators' general observations that without cross referencing the source video, trunks and thick branches in the background could be easily misinterpreted as being part of the foreground plane.

On the Main data set, the Syn+F* network was significantly outperformed by both the RGB+D and RGB+F networks. This is not wholly surprising due to overfitting effects. Due to the manual annotation, the RGB networks also learned to filter out non-branch objects in the foreground, usually ribbons hanging off of the trellis wires, that the Syn networks were not trained to filter out. 

However, Table~\ref{table:pvals} shows that in almost all other scenarios no network was able to significantly outperform Syn+F*, the one exception to this being the RGB+F network on the Afternoon data set. In particular, the Lab environment proved to be extremely challenging for all networks besides the Syn+F* network. Overall, the RGB+F network proved to be the strongest competitor to Syn+F*, outperforming Syn+F* in the Main and Afternoon data sets, matching it in the Cloudy and Envy data sets, and having the second best (though still highly inaccurate) performance in the Lab setup. This provides support that optical flow is, itself, a powerful alternative to depth for foreground segmentation. However, the performance degradations for RGB+F network are more significant than the degradations for the Syn+F* network. This shows that the consistent performance of the Syn+F* network derives not just from its use of optical flow but also from environmental robustness gained from learning using synthetic data. Meanwhile, the RGB+D network showed very significant degradations in performance for every environment, indicating poor robustness to environments differing from the training set.

We acknowledge some shortcomings in our data collection process that could potentially have improved the results of the analysis. First, when obtaining the depth images, we did not align them with the color image frame, meaning that in order to make use of the depth data the pix2pix network would have to learn to do the depth/color alignment itself. Based on the performance of RGB+D versus RGB, the misaligned depth data still appears to be useful, but it is likely that proper depth alignment would have improved the RGB+D performance. However, we note that, from a qualitative examination of the depth images, they were quite noisy, contained many holes and shadowing, and often completely failed to capture thin structures, problems that optical flow had less frequently. Therefore, we do not believe that correcting the depth data would allow RGB+D to perform better than RGB+F in any scenario. 

Also, we would have liked to obtain more training data from different days and conditions to examine the impact of adding more data on a trained network's robustness, which we were unable to do due to time and labor constraints. However, we have already shown that training on RGB data almost always results in reduced robustness when moving to less familiar environments, and it is unlikely that any amount of data collected from outdoor orchards could have allowed an RGB-trained system to function in the Lab environment.

\section{Conclusion}

In this paper, we propose a system for performing branch segmentation in complicated orchard environments in which the foreground and background cannot be easily discerned from a single RGB image alone. In lieu of attempting to control the environment, which is not always feasible, or using depth data, which can be unreliable and requires hand-labeled data to train a neural network, we take advantage of our system design (i.e. an eye-in-hand configuration for robotic pruning). Specifically, we use optical flow in place of depth, allowing our system to be trained using only simulated data with no manual labeling required and to function with any suitable RGB camera. We verify our system's performance against various networks trained on manually-labeled data and prove that the joint use of synthetic data and optical flow data results in a highly robust system that performs well even in challenging, unforeseen environments. 


\section*{ACKNOWLEDGMENT}

The authors would like to acknowledge and thank Olsen Brothers Ranches, Inc. (Prosser, WA) for their support during data collection, and Gopala Krishna Josyula and Abhinav Jain for help with the data labeling process.

\bibliographystyle{IEEEtran}
\bibliography{iros2022.bib}

\addtolength{\textheight}{-12cm}   

\end{document}